\documentclass[10pt,twocolumn,letterpaper]{article}

\usepackage[pagenumbers]{cvpr} % To force page numbers, e.g. for an arXiv version

\usepackage{multirow}
\usepackage[table]{xcolor}

\newcommand{\TODO}[1]{\textbf{\color{red}[TODO: #1]}}
\renewcommand{\TODO}[1]{}

\usepackage{microtype}

\renewcommand{\paragraph}[1]{\vspace{0em}\noindent\textbf{#1.}}

\usepackage{xspace}
\newcommand{\paper}{PhysVid\xspace}

\usepackage{acronym}
\usepackage{algorithm}
\usepackage{algpseudocode}

\definecolor{mypurple}{HTML}{BCBCF6}
\definecolor{myred}{HTML}{F9CCCC}
\definecolor{mygray}{HTML}{BFBFBF}

\acrodef{T2V}{Text-to-Video}
\acrodef{LLM}{Large Language Model}
\acrodef{VLM}{Vision Language Model}
\acrodef{RoPE}{Rotary Positional Embeddings}
\acrodef{GAN}{Generative Adversarial Network}
\acrodef{T2I}{Text-to-Image}
\acrodef{DiT}{Diffusion Transformer}
\acrodef{SA}{Semantic Adherence}
\acrodef{PC}{Physical Commonsense}
\acrodef{CFG}{Classifier-Free Guidance}

\definecolor{cvprblue}{rgb}{0.21,0.49,0.74}
\usepackage[pagebackref,breaklinks,colorlinks,allcolors=cvprblue]{hyperref}

\def\paperID{*****} % *** Enter the Paper ID here
\def\confName{CVPR}
\def\confYear{2026}

\title{\paper: Physics Aware Local Conditioning for Generative Video Models}

\author{
Saurabh Pathak \qquad Elahe Arani \qquad Mykola Pechenizkiy \qquad Bahram Zonooz\\
Eindhoven University of Technology\\
{\tt\small \{s.pathak,e.arani,m.pechenizkiy,b.zonooz\}@tue.nl}\\
[.5em]
{\href{https://5aurabhpathak.github.io/PhysVid}{5aurabhpathak.github.io/PhysVid}}\\
[-.7em]
}
\begin{document}
\twocolumn[{%
\renewcommand\twocolumn[1][]{#1}%
\maketitle
\begin{center}
    \centering
    \captionsetup{type=figure}
    \includegraphics[width=\textwidth,trim=4 4 4 4,clip]{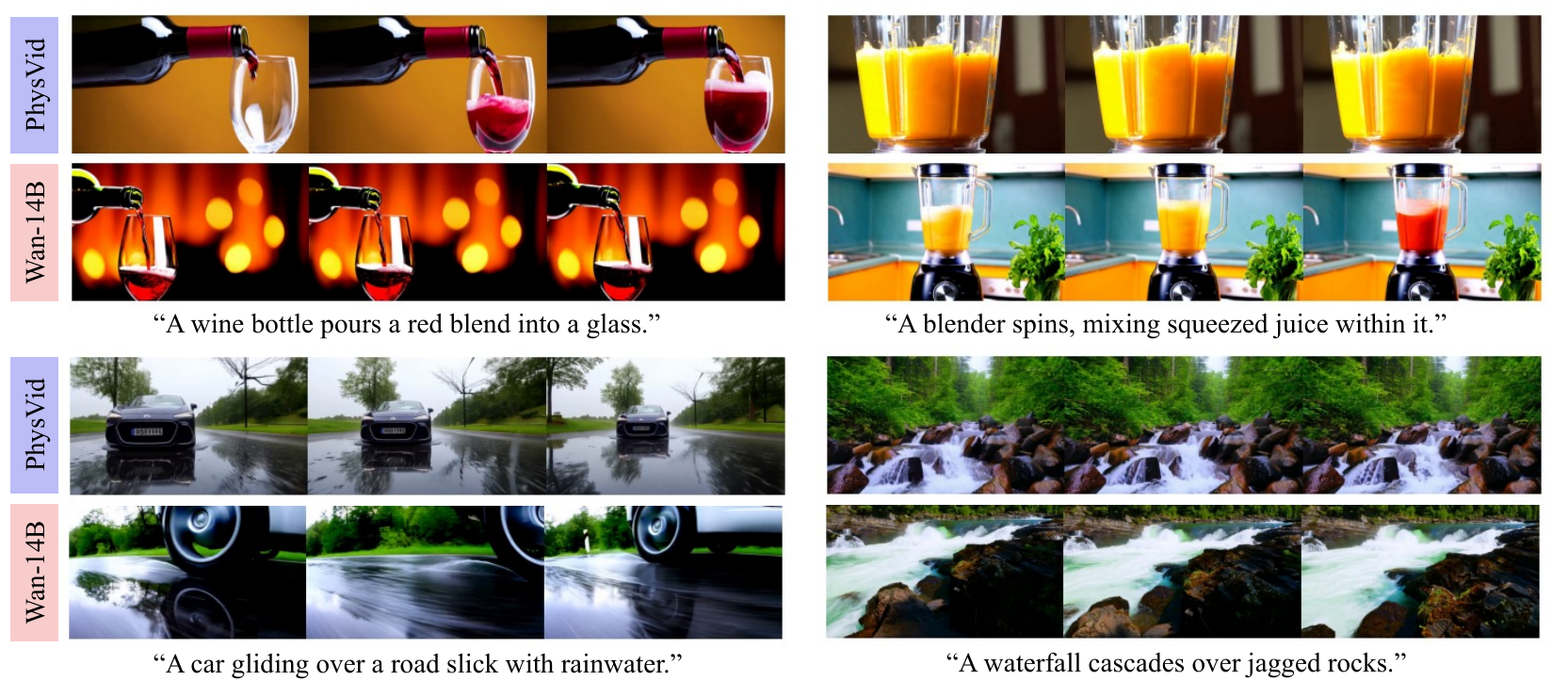}
    \captionof{figure}{Videos generated by \emph{\paper} with \emph{1.7 billion} parameters, compared to videos generated by \emph{Wan-14B}~\cite{wang2025wan} on VideoPhy~\cite{bansal2025videophy} captions. Despite the smaller model size, \emph{\paper} achieves better physical realism in generated videos.}
    \label{fig:main}
\end{center}%
}]
% \maketitle
\begin{abstract}
Generative video models achieve high visual fidelity but often violate basic physical principles, limiting reliability in real‑world settings. Prior attempts to inject physics rely on conditioning: frame‑level signals are domain‑specific and short‑horizon, while global text prompts are coarse and noisy, missing fine‑grained dynamics. We present \paper, a physics‑aware local conditioning scheme that operates over temporally contiguous chunks of frames. Each chunk is annotated with physics‑grounded descriptions of states, interactions, and constraints, which are fused with the global prompt via chunk‑aware cross‑attention during training. At inference, we introduce negative physics prompts (descriptions of locally relevant law violations) to steer generation away from implausible trajectories. On VideoPhy, \paper improves physical commonsense scores by $\approx 33\%$ over baseline video generators, and by up to $\approx 8\%$ on VideoPhy2. These results show that local, physics‑aware guidance substantially increases physical plausibility in generative video and marks a step toward physics‑grounded video models.
\end{abstract}
\section{Introduction}
\label{sec:intro}
Generative video models have seen a remarkable improvement in aesthetic realism and video quality in the past few years, exemplified by successful commercial models such as Sora~\cite{brooks2024video} and Genie~\cite{bruce2024genie}. However, despite being trained on enormous datasets, they still face difficulties in generating videos that faithfully adhere to the physical laws observed in nature and inherent in the data~\cite{kang2025how,motamed2025generative}. This inability points to the challenge and possibly the existence of a fundamental ceiling on learning to generate physically accurate videos from data alone, without any explicit mechanism to incorporate the underlying physics. This problem has been acknowledged in the literature and methods have been proposed to improve the physical accuracy of generated videos by incorporating explicit physics-based constraints or models into the generation process~\cite{gillman2025force,liu2024physgen,wang2025wisa,xue2024phyt2v,yang2025vlipp,zhang2025thinkdiffusellmsguidedphysicsaware,zhang2025videorepa,hao2025enhancing,yuan2023physdiff,yuan2025magictime}.
However, these methods apply physics conditioning at the entire time scale of a video, which limits their ability to capture fine-grained physical phenomena that evolve over shorter time scales. To address this limitation, we aim to discover the physics information that arises at temporally local levels in the data and inject it as an additional sequence-aware conditioning in the generative architecture, distinct from the traditional \ac{T2V} pathway.

The key inspiration to focus on local temporal segments during the video generation process comes from the observation that global text conditioning may be insufficient to capture the intricate physical interactions that occur over subintervals. Previous approaches have focused on enhancing global prompts with physics-based information~\cite{zhang2025thinkdiffusellmsguidedphysicsaware,xue2024phyt2v}. However, doing so does not guarantee that the model will focus on relevant details within the appropriate subinterval of the video being generated. Although effective for static image generation, recent research has shown that global cross-attention mechanism that applies the same textual guidance across all frames can be suboptimal for video generation, as the model may struggle to interpret the temporal logic of the prompt, leading to a failure to generate details specific to local time intervals~\cite{luo2025enhanceavideo,xia2024unictrl,feng2024fancyvideo,zheng2025frame}. This limitation is demonstrated in models where temporally consistent textual guidance results in nearly static attention maps for action-related words over time, causing the generated video to exhibit static or incoherent motion due to the spatiotemporal misalignment of global conditioning with generated frames~\cite{feng2024fancyvideo, tian2024videotetris}. For this reason, the motion of objects, changes in lighting conditions, and interactions between elements in a scene often occur rapidly and can be better described when considered in smaller intervals. By conditioning each temporal segment on relevant physics principles within it, we can ensure that it adheres to physical laws more closely, resulting in a coherent and realistic overall video. We are also inspired by recent progress in video-based world modeling, where video generation is dynamically controlled with frame-level modulations~\cite{gao2024vista,cheng2025playing,yu2025gamefactory,huang2025vid2world,he2025pretrained,jang2025frame,zheng2025frame,hong2024large}. We extend this idea of frame-level control to physics conditioned video generation over short temporal fragments.

The proposed approach \emph{\paper}, involves the following steps: First, the target video is segmented into smaller temporal fragments. Next, the observable physical phenomena in each segment are analyzed, identifying key physical dimensions such as motion dynamics, shape deformations, and optical effects with the help of a \ac{VLM}. This information is used to annotate each segment with a corresponding physics-aware prompt to directly support its content during generation. Finally, we train a video generation model with temporally aware cross-attention layers that incorporate the segment-level physics-based prompts alongside the global text prompt. This allows the model to respond to both the global context and local physical phenomena during generation. We validate our approach through extensive experiments on the WISA-80k dataset~\cite{wang2025wisa}.

In summary, we propose the following key contributions.

\begin{itemize}
    \item We incorporate additional text conditioning pathways into a \ac{T2V} generator. In contrast to frame-level action conditioning and global text conditioning, our method acts on groups of frames. Working at the chunk~\footnote{We use the term ``chunk'' to refer to a temporally contiguous set of frames from a video.} level preserves sufficient temporal information necessary to observe physical laws locally, such as motion, while avoiding locally irrelevant pieces of information from the global text.
    \item We create a separate text prompt for each chunk using a \ac{VLM}. During generation, each chunk is supported by its own physics based text conditioning in addition to the global \ac{T2V} prompt.
    \item During inference, we also generate counterfactual prompts for each chunk based on the violation of locally observable physics laws. We use these prompts to guide the video generation away from the physically implausible scenarios.
\end{itemize}

In the following sections, we first describe the background, followed by a description of \paper. We then report the results of our experiments and conclude with a discussion section.

\section{Background}
\label{sec:background}

\subsection{Generative Text-to-Video modeling}
Several methods to generate videos from text description have been proposed in the literature. Earlier methods that laid the foundation for conditional video generation were based on \acp{GAN}~\cite{deng2019irc,kim2020tivgan,li2018video}. However, these approaches suffered from training instability and temporal consistency. Later work shifted towards transformer based autoregressive \ac{T2V} generation~\cite{hong2023cogvideo,wu2022nuwa,wu2021godiva,villegas2023phenaki}. These models, characterized by sequential prediction of discrete video tokens, can generate videos faster than their predecessors. However, they are susceptible to rapid accumulation of errors, leading to degradation of temporal coherence over long sequences.
In contrast, diffusion based methods avoid the need for discrete tokenization by operating in continuous spaces. A large body of work has adapted existing \ac{T2I} architectures for video tasks, bypassing the need for massive text-video datasets~\cite{khachatryan2023text2video,guo2024animatediff,singer2023makeavideo,blattmann2023align,blattmann2023stable,an2023latent,wang2023modelscope,yuan2024inflation,wu2023tune,zhou2022magicvideo,wang2025lavie,bar2024lumiere}. This is achieved by modifying the internal mechanisms of \ac{T2I} models, such as by augmenting with additional temporal layers, structuring latents, or attention techniques to create temporal consistency.
With increased availability of large \ac{T2V} datasets such as OpenVid~\cite{nan2025openvidm}, WebVid~\cite{bain2021frozen} and Panda~\cite{chen2024panda}, a growing line of work has focused on spatiotemporal diffusion by directly modeling videos in 4D pixel or latent spaces, thus denoising all frames jointly without relying on image based spaces or \ac{T2I} backbones~\cite{ho2022video,ho2022imagen,qin2024xgen,yang2025cogvideox,menapace2024snap}.
Beyond diffusion, recent work has begun to explore flow‑matching techniques~\cite{lipman2023flow} for video generation, requiring much fewer generation steps than diffusion based models~\cite{jin2025pyramidal,cao2025video,polyak2024moviegen,wang2025wan,chen2025goku}.
In this context, we consider a \ac{T2V} generative model $\mathcal{G}$ that iteratively transforms over $T$ time steps, a 4D Gaussian noise sample $x_T \sim \mathcal{N}(0,\mathcal{I}) \in \mathbb R^{F\times C\times H\times W}$ under a conditional text prompt $c$, into a video $x_0$ that follows the content described in $c$.

\paragraph{Cross attention in \ac{T2V} modeling} Architecturally, \ac{T2V} models have progressed from 3D U‑Net backbones~\cite{ho2022video} that denoise spatiotemporal volumes, to \acp{DiT}~\cite{peebles2023scalable} that scale more effectively and capture longer‑range dependencies with space–time self-attention. The text conditioning in \ac{DiT} benefits significantly from its cross-attention mechanism, where the visual features of the noisy latent video act as the `query', while the embeddings from a text encoder serve as the `key' and `value'. This allows the model to dynamically weigh the importance of different parts of the text prompt for different spatiotemporal locations in the generated video. However, in prevalent \ac{T2V} pipelines, cross-attention operates globally. Every spatiotemporal token in the video latent attends to the same time‑agnostic text tokens, thus textual guidance is applied across frames without frame specific conditioning. This approach is inefficient across temporal semantics that change over time, motivating our approach that ties different captions to temporal segments for improved temporal alignment.

\subsection{Physics-aware video generation}
A recent line of research has focused on explicitly incorporating physical principles into the video generation process. Existing approaches include the use of physics simulators~\cite{gillman2025force, liu2024physgen}, the incorporation of physical constraints into loss functions during training~\cite{wang2025wisa, zhang2025thinkdiffusellmsguidedphysicsaware,zhang2025videorepa}, as guidance mechanisms during inference~\cite{hao2025enhancing, yuan2023physdiff}, or the use of modular approaches that incorporate physical awareness into visual generation through multistage generation processes or specialized modules~\cite{wang2025wisa,liu2024physgen,xue2024phyt2v,zhang2025thinkdiffusellmsguidedphysicsaware,yang2025vlipp}.
``Force Prompting''~\cite{gillman2025force} is a technique in which video generation is conditioned on explicit physical forces. This method allows a user to apply localized or global forces, such as pushes or winds, to an initial image. The model then generates a video sequence in which objects react according to these physical inputs, enabling a form of interactive and physically responsive video synthesis.
DiffPhy~\cite{zhang2025thinkdiffusellmsguidedphysicsaware} employs a \ac{LLM} to analyze a text prompt and infer its underlying physical context, such as gravity, collisions, or momentum. The \ac{LLM} generates an enhanced, physics-aware prompt that provides explicit guidance to the video diffusion model. To ensure that the final output adheres to these principles, a multimodal \ac{LLM} acts as a supervisor, evaluating the physical correctness of the generated frames, and guiding the model's training process.
PhyT2V~\cite{xue2024phyt2v} is a training-free method that improves the physical realism of videos generated through multiple rounds of generation and refinement. In this method, an initial \ac{T2V} prompt is used to generate a video which is then captioned. An \ac{LLM} then refines the prompt for the subsequent round based on mismatches between the caption and the current prompt. This approach eventually improves the physical plausibility of the generated video, but requires several rounds of generation.
\citet{hao2025enhancing} recently proposed a training-free approach that uses \ac{LLM} to reason about the governing physical principles corresponding to a \ac{T2V} prompt and generates a counterfactual prompt that violates them. During inference, both the original and the counterfactual prompts are used in a guidance based generation mechanism similar to \ac{CFG}~\cite{ho2021classifierfree}.
WISA~\cite{wang2025wisa} decomposes abstract physical principles into textual descriptions, qualitative categories, and quantitative properties, and injects them via a mixture of physical experts and a physical classifier. It also curates a dataset that covers diverse laws in dynamics, thermodynamics, and optics to train and evaluate physics compliance.
VideoREPA~\cite{zhang2025videorepa} transfers physics understanding from video foundation models to \ac{T2V} models using cross distillation losses, aligning intraframe spatial and interframe temporal relations to improve physical commonsense without relying on physics specialized datasets.

\begin{figure}
  \centering
    \includegraphics[width=\columnwidth]{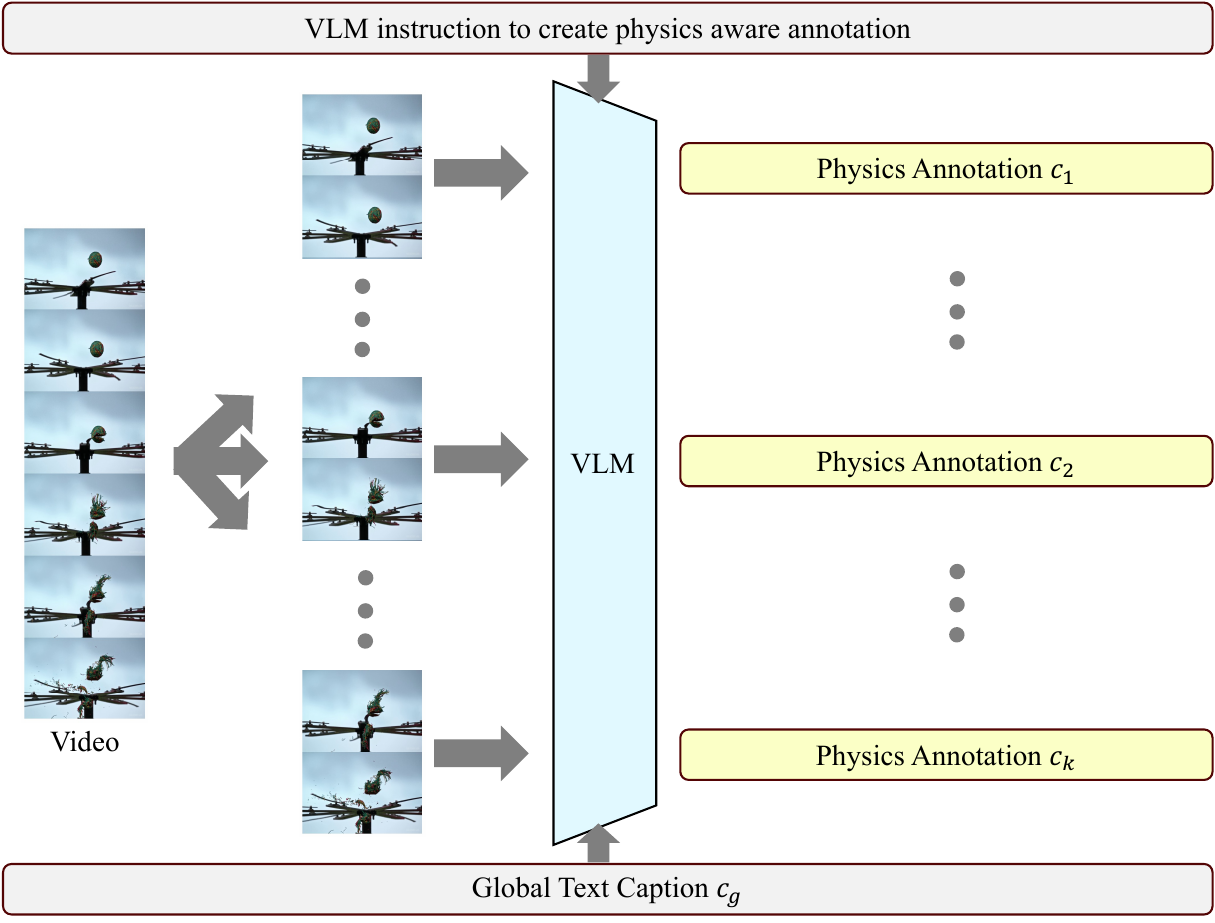}
    \caption{Procedure for generating physics-grounded local prompts during data annotation.}
    \label{fig:data_annotation}
\end{figure}

\begin{figure*}
  \centering
    \includegraphics[width=.7\textwidth]{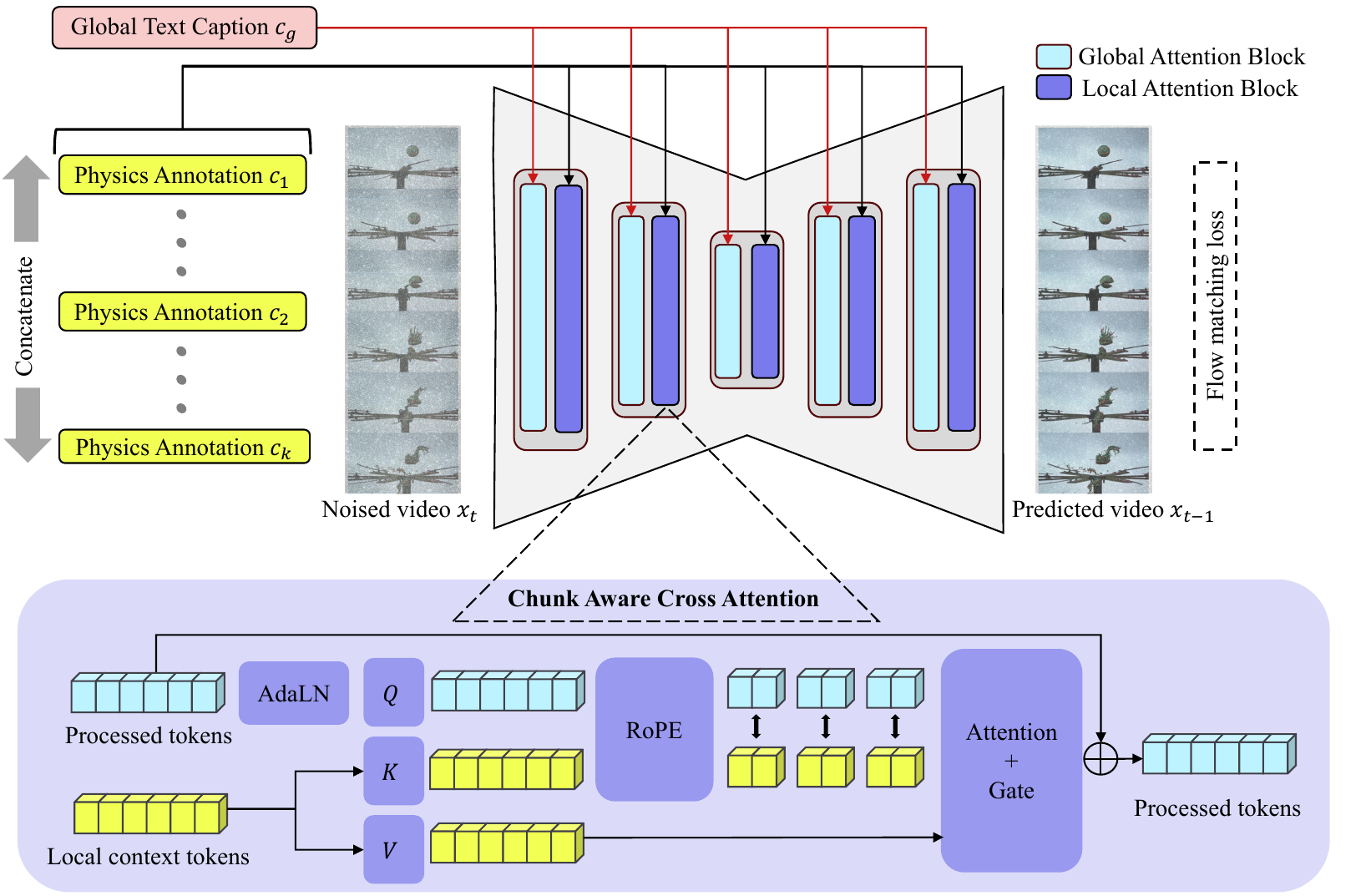}
    \caption{Architecture of \emph{\paper} showing local information pathways with chunk aware cross-attention. Commonly applied procedures such as tokenization, latent encoding, and decoding are implicit and not shown.}
    \label{fig:training}
\end{figure*}

\section{Method}
\label{sec:method}
The central objective of the proposed approach is to improve the overall quality of observable physical phenomena in the generated videos. To that end, we incorporate additional text conditioning based on local physical phenomena observed within smaller temporal segments of the video. This local conditioning is used in conjunction with global \ac{T2V} conditioning to enhance the physical realism of the generated videos. Specifically, given a global prompt $c_g$, and a set of $k$ local prompts $C:=\{c_1, \ldots, c_k\}$ based on physics and aligned with $c_g$, our physics-aware generative video model $\mathcal G$ generates a video $x_0$ grounded in both $c_g$ and $C$, by iterative denoising over $T$ steps.
\begin{align}
    x_{T-1} = \mathcal{G}(x_T, c_g, C, T) \qquad x_T \sim \mathcal N(0, I)
    \label{eq:main}
\end{align}

In this section, we first describe the procedure for generating physics-based annotations for local conditioning. Subsequently, we describe our chunk-aware cross-attention mechanism that powers the local conditioning pathway, inserted as additional layers in the base architecture. Lastly, we explain the inference process that involves counterfactual generation to enable guidance.

\subsection{Annotation of video chunks}
\label{sec:videoannotation}
Given a training dataset of videos paired with text captions, we first annotate each video with chunk-level physics-based prompts. To do this, we divide each video in the training data set into a set of contiguous fixed-duration temporal chunks, with a fixed number of frames comprising each chunk. As shown in~\cref{fig:data_annotation}, each chunk is then separately analyzed by a \ac{VLM} to identify the visible elements and physical phenomena contained within that segment.
While analyzing each chunk separately in this manner helps to focus on the local information, it risks the generated annotations becoming misaligned with the global caption or even directly contradicting it in the worst case. We address this by also providing the global \ac{T2V} prompt to the \ac{VLM} as part of its instructions when processing each chunk and encourage it to align its annotations with the visible content without contradicting the global \ac{T2V} prompt.
The \ac{VLM} generates a structured description of physical phenomena within the chunk, focusing on visible physical phenomena. We instruct the \ac{VLM} to focus on three key categories of physics information: dynamics, shape, and optics. The instructions provided to the \ac{VLM} encourage it to work step by step in a structured manner. This ensures that the descriptions are physically accurate and relevant to the video chunk. In addition, we apply constrained generation techniques to strictly enforce this structure in the output~\cite{willard2023efficient}. The structured output from the \ac{VLM} is parsed to extract the relevant physics information, which is then converted into a concise text prompt corresponding to that video chunk. This prompt is used as local conditioning for that chunk during the training. An example of the prompt used to guide the \ac{VLM} is provided in Appendix~\ref{appendix:posprompt}.

\subsection{Local conditioning with cross-attention}
Given the annotated dataset with chunk-level physics-based prompts, we train a model to incorporate this local conditioning. The general architecture of the model is illustrated in~\cref{fig:training} and the procedure is described in~\cref{alg:ropealign}. Specifically, we employ chunk-aware local cross-attention in which \ac{RoPE}~\cite{su2024roformer} is applied to both vision and text modalities. Similarly to standard self-attention, video query tokens are modulated by \ac{RoPE} parameterized by the 3D spatiotemporal grid (frame, height, width). However, we also apply \ac{RoPE} to text key projections with an identical frequency basis for both video query and text key projections. To track positional awareness of chunks within the local textual information flow, we define a text grid that includes a chunk axis aligned to the number of video chunks. In this manner, cross-attention logits become explicitly cross-modal position aware. It enables a video token to attend differently to text information from a different chunk compared to text assigned to its own chunk. This design contrasts with conventional text‑to‑video cross‑attention, in which only video queries carry video positional encoding, and text keys follow one‑dimensional textual positions, thereby lacking frame‑aligned coupling. These modules are then inserted into a pretrained model inside each transformer block and trained with flow matching~\cite{lipman2023flow}. This chunk-wise design enforces local temporal neighborhoods, while a parallel global cross-attention path preserves long-range conditioning.
In general, this design promotes temporally grounded text-video alignment while remaining compatible with standard \ac{T2V} architectures.

\begin{figure}
  \centering
  \includegraphics[width=\columnwidth]{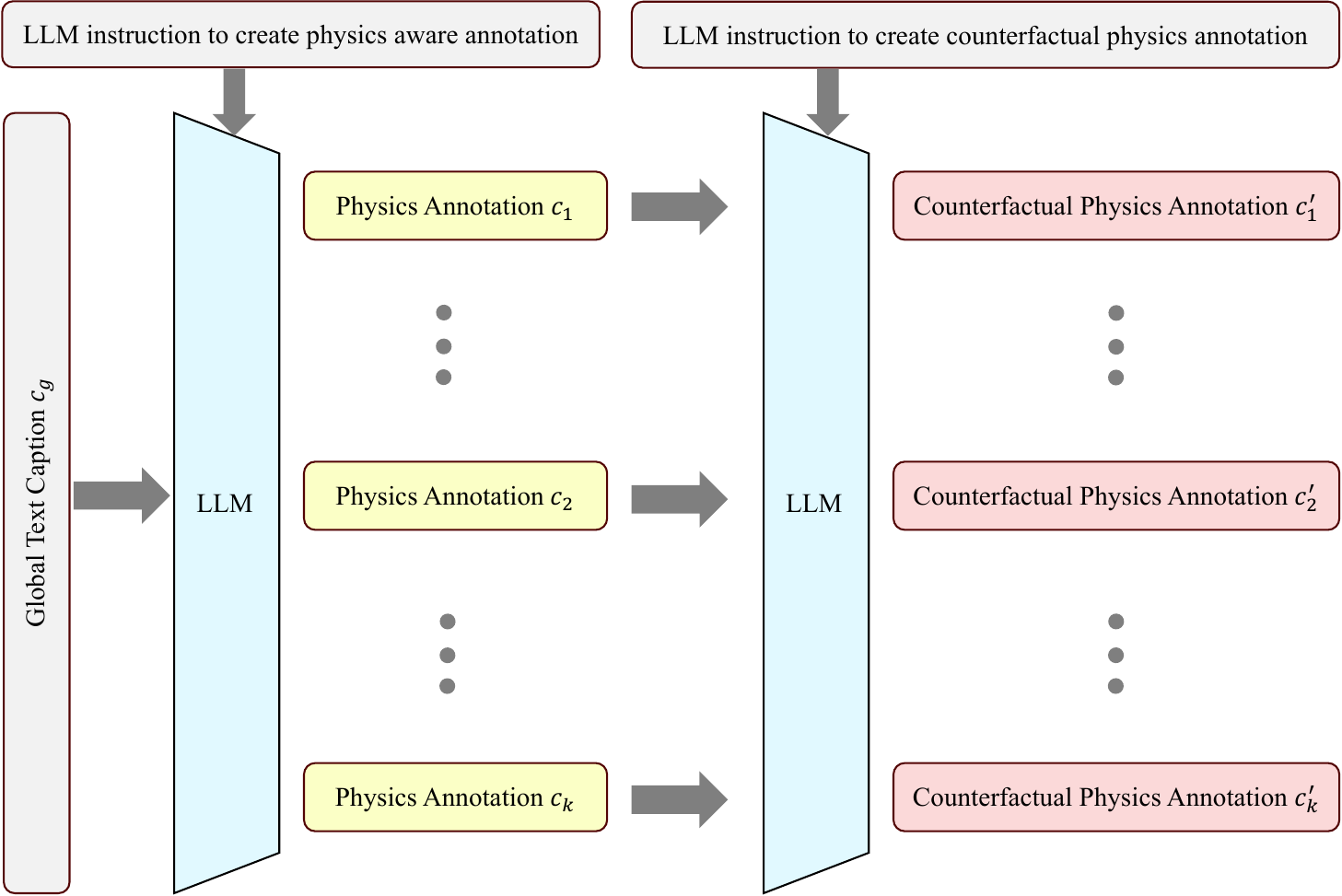}
    \caption{Generation of local and counterfactual prompts during inference.}
    \label{fig:neg_annotation}
\end{figure}
\subsection{Inference with local counterfactual guidance}
During inference, only the global \ac{T2V} captions are available, since the videos must be generated from pure noise.
Therefore, to supply the local conditioning pathways in the model, we generate the local captions from only the global text prompt, as shown in~\cref{fig:neg_annotation}. Here, we instruct a \ac{LLM} to generate a set of prompts grounded in local physics for an ``imagined'' video clip using only the information present in the global text. These prompts are required to be temporally coherent. In addition, a local prompt is allowed to contain information that is not mentioned in the global prompt, as long as it does not break alignment with the global prompt or the preceding local prompts. It is important to note that during inference, the annotations corresponding to all chunks in a video are generated together, in contrast to the training data annotation process explained in~\cref{sec:videoannotation}. This is done because of the lack of local visual data during inference and also to reduce overhead. The instruction used for \ac{LLM} to generate these prompts is provided in Appendix~\ref{appendix:infprompt}.

\paragraph{Counterfactual generation} In addition to generating a prompt that accurately describes the physical phenomena in each chunk, we also generate a corresponding counterfactual that deliberately violates those phenomena using a process similar to~\citet{hao2025enhancing}. To create a counterfactual prompt, an \ac{LLM} first identifies key visual and physics-relevant elements in the ``original'' local prompt generated previously. It then generates a counterfactual statement that directly contradicts these physics observations, while still being relevant to the visual elements in the original. The generated counterfactual prompts are used during the inference stage to guide the model away from generating physically inaccurate content. The instruction used to generate the counterfactual prompt is shown in the Appendix~\ref{appendix:negprompt}.

During inference, we use both positive and counterfactual local prompts to guide the generation process. We employ classifier-free guidance~\cite{ho2021classifierfree} at both global and chunk levels, as follows.
\begin{align}
    x_{T-1} =& (1+w)\cdot\mathcal{G}(x_T, c_g, C, T)-\\ \nonumber
    &w\cdot\mathcal{G}(x_T, c_n, C', T),
    \label{eq:cfg}
\end{align}
where $c_n$ is a fixed global negative prompt similar to Wan~\cite{wang2025wan}.
$C'$ is the set of counterfactual prompts paired with the set of corresponding physics-based prompts $C$, and $w$ is the guidance scale. Other terms are similar to~\cref{eq:main}.
This strategy enhances the physical accuracy of the generated videos by reinforcing correct physics while discouraging incorrect representations, effectively steering the model away from generating content that violates physical laws.

In the next section, we demonstrate that, equipped with the attributes described in this section, \paper is capable of inducing adherence to physical principles in the synthesized video content.
\section{Experiments}
\label{sec:experiments}
We begin this section with an explanation of our setup and data preparation procedure and the choice of evaluation benchmarks. Subsequently, we present quantitative and qualitative results on the evaluation benchmarks, comparing our method with the baselines and the previous literature. Finally, we present the results of the ablation experiments that analyze the impact of our method with and without the aid of counterfactual guidance during generation and compare it with plain finetuning.

\subsection{Setup}
\paragraph{Data} We use the WISA~\cite{wang2025wisa} data set that contains a diverse collection of more than $80$ thousand videos related to various physical phenomena observed in the world. Following the configuration in Wan~\cite{wang2025wan}, we remove videos less than $5$ seconds long and divide the remaining videos into $5$ second clips sampled at $832\times 480$ at $16$ frames per second, leading to a total of $81$ frames per video. This results in approximately $53$ thousand video samples. Note that while WISA provides detailed physics annotations for each of their training samples, we do not utilize them in our approach, relying instead on learning this information purely from the training videos. This is because these annotations are at a global level and do not focus on physical phenomena observed within smaller temporal segments of the video. Furthermore, due to the division of long videos into $5$ second clips, these global annotations become misaligned with the video clips and may lead to noisy conditioning if used directly. This strategy also has the added benefit of making our approach generalizable to other datasets that do not contain such explicit information about physical phenomena.
Subsequently, we generate our own chunk-level physics-based prompts using the method described in Section~\ref{sec:method} as well as a global text caption for the entire video. We use VideoLLama3-7B~\cite{zhang2025videollama3} for this task. First, we generate a global caption for the entire $5$ second video clip. Thereafter, the video is divided into $7$ temporally contiguous chunks of frames of approximately $0.7$ seconds each. Next, we generate a segment-level prompt using the video chunk along with the global caption as input to the \ac{VLM}, processing each chunk separately in this manner. Examples of the generated prompts are provided in Appendix~\ref{appendix:example_prompts}.

\paragraph{Model} We introduce chunk-aware cross-attention layers in each transformer block in a pretrained Wan2.1~\cite{wang2025wan} model with $1.3$ billion-base parameters.
The entire architecture is trained in two stages. First, the base model is frozen, and only the newly added modules are trained for $1000$ steps. Once they have stabilized, the base layers are unfrozen, and the entire model is further trained for additional $2000$ steps.
We use $4$ GPUs for this task, leading to an effective batch size of $64$ samples per step. To generate annotations during inference as discussed in~\cref{sec:method}, we utilize VideoLLama3-7B as a language model.

\subsection{Evaluation benchmarks}
We use two recently proposed and widely adopted benchmarks, VideoPhy~\cite{bansal2025videophy} and VideoPhy2~\cite{bansal2025videophy2corr}, to evaluate the physical accuracy of the generated videos. VideoPhy consists of $344$ manually curated captions in three different categories of physical interactions, namely, ``solid-solid'', ``solid-fluid'', and ``fluid-fluid''. 
Similarly, VideoPhy2 uses a larger test set with $590$ captions providing coverage over a diverse set of real-world physical phenomena. Their data is divided into two main action categories: ``object interactions'' and ``sports and physical activities''.
Both these benchmarks include an additional category comprising of manually labeled ``hard'' examples. In addition, they provide an automatic evaluation model that is trained on human annotations of the generated videos. As a result, their evaluation scores are correlated with human judgment of physical correctness. We follow the same scoring mechanism as outlined in the respective original works. All results are reported as the mean of the benchmark scores for $5$ different evaluation sets generated with different random seeds. We present the results next.

\subsection{Results}
\subsubsection{Quantitative results}
\label{sec:quantresults}
\begin{figure}
  \centering
  \includegraphics[width=.8\columnwidth,trim=5 9 8 10,clip]{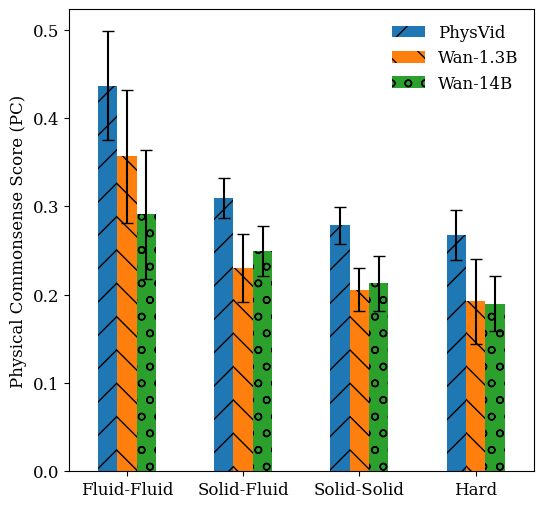}
    \caption{VideoPhy \acf{PC} score by category.}
    \label{fig:videophy}
\end{figure}

\paragraph{VideoPhy} \Cref{tab:videophy12} shows the performance of \paper on the VideoPhy benchmark compared to two Wan2.1 baselines. With a model size of only 1.7B parameters, \paper significantly outperforms both the smaller (1.3B) and the much larger (14B) base model in the physical commonsense metric by $\approx 33\%$. Furthermore, this performance gain is reflected across all subcategories in the benchmark as shown in~\cref{fig:videophy}, demonstrating the effectiveness of local information in improving physical awareness of the generative video model.

\begin{table}
  \caption{Results on VideoPhy and VideoPhy2. We report semantic alignment (SA) and physical commonsense (PC; higher is better). \paper\ (1.7B) achieves the best PC on both benchmarks ($\approx 33\%$ relative gain on VideoPhy and over $8\%$ on VideoPhy2 vs. Wan-14B).}
  \label{tab:videophy12}
  \centering
  \begin{tabular}{@{}lccccl@{}}
    \toprule
    \multirow{2}{*}{Method} & \multirow{2}{*}{Params (B)} & \multicolumn{2}{c}{VideoPhy} & \multicolumn{2}{c}{VideoPhy2} \\ 
    \cmidrule(lr){3-4} \cmidrule(lr){5-6}
    & & SA & PC & SA & PC \\ \midrule
    Wan-1.3B & $1.3$ & $0.46$ & $0.24$ & $0.28$ & $0.61$ \\
    Wan-14B & $\mathbf{14}$ & $\mathbf{0.52}$ & $0.24$ & $\mathbf{0.29}$ & $0.59$ \\
    \paper & $1.7$ & $0.43$ & $\mathbf{0.32}$ & $0.28$ & $\mathbf{0.64}$ \\
    \bottomrule
  \end{tabular}
\end{table}

\begin{figure}
  \centering
  \includegraphics[width=.8\columnwidth,trim=5 9 8 10,clip]{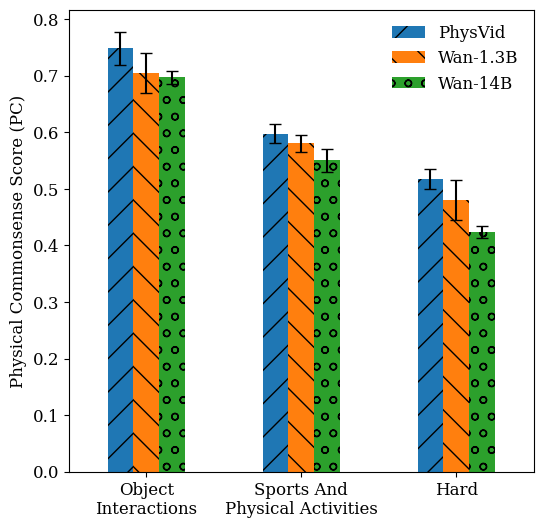}
    \caption{VideoPhy2 \acf{PC} score by category.}
    \label{fig:videophy2}
\end{figure}

\paragraph{VideoPhy2} As shown in~\cref{tab:videophy12}, \paper performs, respectively, $\approx 5\%$ and $\approx 8\%$ better than the corresponding baselines on the physical commonsense score. As~\cref{fig:videophy2} shows, this improvement is consistent across both ``object interaction'' and ``sports and physical activities'' subcategories of the benchmark, as well as on the captions that are categorized by the benchmark as hard to generate accurately. Compared to VideoPhy, the improvements of our method are significantly less pronounced on VideoPhy2. The reason behind this could be the difference in the score calculation method between the two evaluation approaches. Specifically, VideoPhy2 uses a categorical rating system in contrast to VideoPhy, which uses a rating on a continuous scale of $0$ and $1$, followed by hard thresholding~\cite{bansal2025videophy, wang2025wisa}.

\paragraph{Comparisons with existing approaches}~\Cref{tab:comparisons} shows the performance of existing methods on the widely adopted VideoPhy benchmark. We collate results from previous work and ours. From this table, it can be observed that the physical accuracy of generative video models does not necessarily scale up with the model size. Furthermore, the performance of a generative video model may vary significantly across multiple evaluations. Although this could be due to differences in underlying settings (\eg, number of denoising steps), it is important to remember that the VideoPhy auto-evaluator is trained to imitate human judgment on the generated videos, which is subjective in nature and inherently noisy. Nevertheless, existing results show that with just a $1.7$ billion total parameters, \paper remains competitive on the VideoPhy benchmark and is on par with the current state-of-the-art, matching or even surpassing many of the larger models.

\begin{table}
  \caption{Quantitative comparisons of our approach with previous works evaluated on the VideoPhy benchmark, including both general and physics-aware generative video methods. Symbols indicate results taken from prior work: WISA~\cite{wang2025wisa} ($\dagger$),VideoREPA~\cite{zhang2025videorepa} ($\$$),~\citet{hao2025enhancing} ($\#$),PhyT2V~\cite{xue2024phyt2v} ($\sim$),VideoPhy~\cite{bansal2025videophy} ($\ddagger$). Our own baselines are marked with $\ast$. Best model scores are underlined. For physics-aware methods with multiple entries, we report the best \ac{PC} score, with parentheses showing relative improvement over the corresponding baseline.}
  \label{tab:comparisons}
  \centering
  \begin{tabular}{@{}lcc@{}}
    \toprule
    % Method & \multicolumn{2}{c}{VideoPhy} \\
    % & SA & PC\\
    Method & SA & PC\\
    \midrule
    Lavie$^\ddagger$ & $0.49$ & $0.28$ \\
    Lavie$^\$$ & $0.49$ & $\underline{0.32}$ \\
    \midrule
    VideoCrafter2$^\dagger$ & $0.47$ & $\underline{0.36}$ \\
    VideoCrafter2$^\ddagger$ & $0.49$ & $0.35$ \\
    VideoCrafter2$^\$$ & $\underline{0.50}$ & $0.30$ \\
    VideoCrafter2$^\sim$ & $0.24$ & $0.15$ \\
    \midrule
    OpenSora$^\#$ & $\underline{0.38}$ & $\underline{0.43}$ \\
    OpenSora$^\ddagger$ & $0.18$ & $0.24$ \\
    OpenSora$^\sim$ & $0.29$ & $0.17$ \\
    \midrule
    HunYuanVideo$^\dagger$ & $0.46$ & $0.28$ \\
    HunYuanVideo$^\$$ & $\underline{0.60}$ & $0.28$ \\
    \midrule
    Cosmos-7B$^\dagger$ & $\underline{0.57}$ & $0.18$ \\
    Cosmos-7B$^\#$ & $0.52$ & $\underline{0.27}$ \\
    \midrule
    CogVideoX-2B$^\ddagger$ & $0.47$ & $\underline{0.34}$ \\
    CogVideoX-2B$^\$$ & $\underline{0.52}$ & $0.26$ \\
    CogVideoX-2B$^\sim$ & $0.22$ & $0.13$ \\
    \midrule
    CogVideoX-5B$^\dagger$ & $0.60$ & $0.33$ \\
    CogVideoX-5B$^\ddagger$ & $\underline{0.63}$ & $\underline{0.53}$ \\
    CogVideoX-5B$^\$$ & $0.63$ & $0.31$ \\
    CogVideoX-5B$^\#$ &$0.48$ & $0.39$ \\
    CogVideoX-5B$^\sim$ &$0.48$ & $0.26$ \\
    \midrule
    Wan2.1-1.3B$^\ast$ & $0.46$ & $0.24$ \\
    \midrule
    Wan2.1-14B$^\#$ & $0.49$ & $\underline{0.35}$ \\
    Wan2.1-14B$^\ast$ & $\underline{0.52}$ & $0.24$ \\
    \bottomrule
    \rowcolor{gray!20} Physics-aware approaches & &\\
    \midrule
    PhyT2V~\cite{xue2024phyt2v} & $0.59(+23\%)$ & $\underline{0.42}(+62\%)$ \\
    PhyT2V$^\dagger$ & $\underline{0.61}(+2\%)$ & $0.37(+12\%)$ \\
    \midrule
    WISA~\cite{wang2025wisa} & $0.67(+12\%)$ & $0.38(+15\%)$ \\
    VideoREPA-5B~\cite{zhang2025videorepa} & $0.72(+14\%)$ & $0.40(+29\%)$ \\
    \citet{hao2025enhancing} & $0.49(+0\%)$ & $0.40(+14\%)$ \\
    \textbf{\paper-1.7B} & $0.43(-7\%)$ & $0.32(+33\%)$  \\
    \bottomrule
  \end{tabular}
\end{table}

\begin{table*}[t]
  \caption{Ablations. The highest scores in each metric are highlighted, whereas the lowest scores are underlined.}
  \label{tab:ablations}
  \centering
  \begin{tabular}{@{}lccccccl@{}}
    \toprule
    \multirow{2}{*}{Method} & \multicolumn{2}{c}{VideoPhy} & \multicolumn{2}{c}{VideoPhy2} \\ 
    \cmidrule(lr){2-3} \cmidrule(lr){4-5}
    & SA & PC & SA & PC\\
    \midrule
    baseline (Wan-1.3B) & $\mathbf{0.4570}$ & $\underline{0.2401}$ & $\mathbf{0.2845}$ & $\underline{0.6144}$ \\
    fine tuning & $\underline{0.4174}$ & $0.2866$ & $\underline{0.2765}$ & $0.6261$ \\
    \paper w/o counterfactual guidance & $0.4355$ & $0.2924$ & $0.2791$ & $0.6334$ \\
    \paper & $0.4302$ & $\mathbf{0.3169}$ & $0.2775$ & $\mathbf{0.6411}$  \\
    \bottomrule
  \end{tabular}
\end{table*}

\begin{figure*}[t]
  \centering
  \includegraphics[width=\textwidth,trim=4 8 4 4,clip]{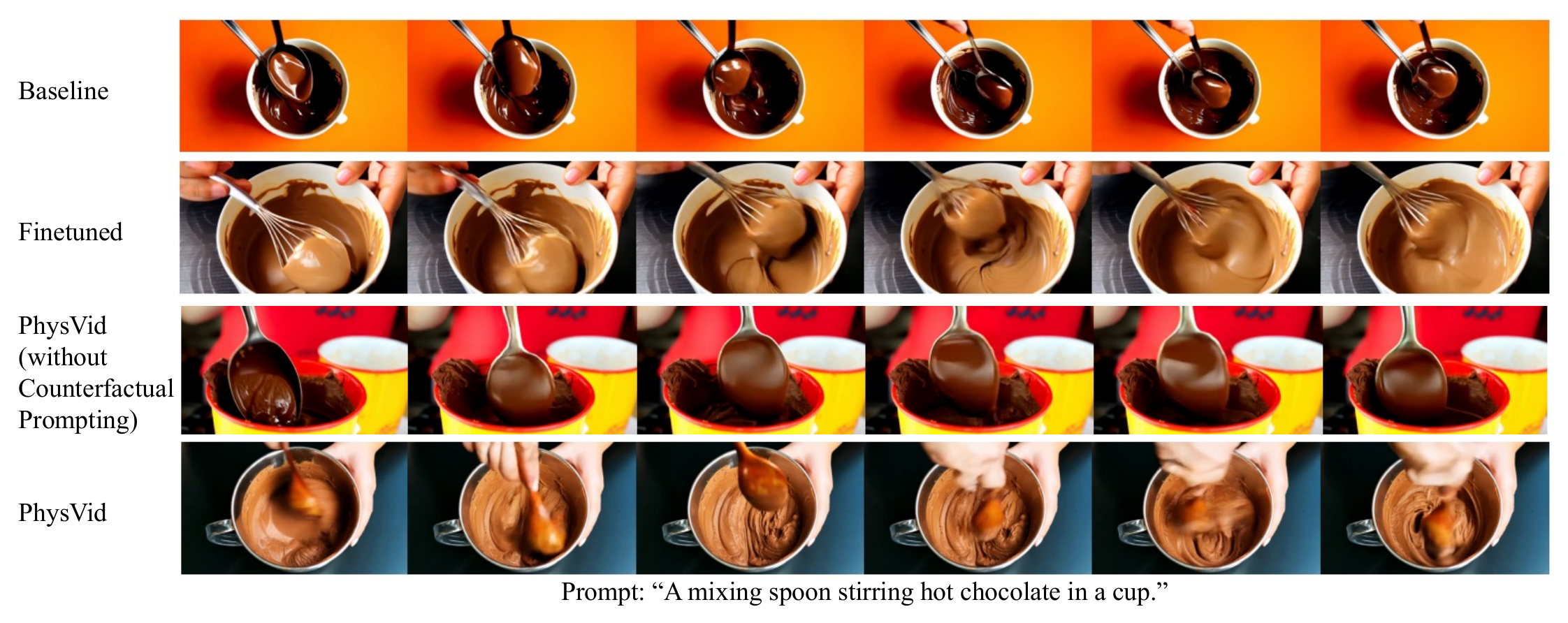}
    \caption{Qualitative ablation example on a VideoPhy prompt. Best viewed when zoomed in.}
    \label{fig:qualitative_ablation}
\end{figure*}

\subsubsection{Qualitative results}
\Cref{fig:main} shows qualitative results on VideoPhy examples. Videos generated with \paper show a visible improvement in the physical fidelity of the content, in contrast to more than $8\times$ larger Wan-14B model. Additional qualitative results are available in Appendix~\ref{appendix:additional_qualitative}.

\subsubsection{Ablations}
\label{sec:ablations}
We perform analysis on the effectiveness of applying locally grounded physics-based text conditioning in generative video modeling in contrast to learning physics information purely from data and standard text conditioning. To that end, we finetune 1.3B baseline with the same dataset but without adding any chunk-aware cross attention layers. Furthermore, to understand the role of counterfactual physics guidance, we also tested the performance of our model both with and without using counterfactual guidance during inference. When inference is performed without using counterfactual prompts, we apply blank local prompts to all local chunks during classifier-free guidance. We evaluate all the approaches on both VideoPhy and VideoPhy2. It is evident from the results in~\cref{tab:ablations} that the use of our method provides clear advantages over pure finetuning. In addition, applying negative physics conditioning to local pathways during guidance based generation further improves the \acf{PC} score.
This is also reflected in our qualitative analysis in~\cref{fig:qualitative_ablation}.

\section{Discussion}
\label{sec:conclusion}
We have presented \emph{\paper}, a method to improve awareness of physical phenomena in generative video modeling by injecting physics knowledge into text prompts aligned with the local temporal segments of a video being generated. We extract this information from videos in the training data using a pretrained \ac{VLM}. The \ac{VLM} is instructed to annotate each chunk of frames with the relevant physics information visible within the chunk. This information is injected into a pretrained generative video model with the help of additional cross attention blocks that employ \ac{RoPE} to align each annotation with its corresponding chunk. We evaluate this approach on VideoPhy and VideoPhy2, two of the widely adopted benchmarks for the evaluation of physical plausibility in generative \ac{T2V} models. The results show a clear improvement in the physical fidelity of the videos generated across both benchmarks. Although we employ a context-specific dataset with labeled physics information, our approach instead relies on its ability to extract this information directly from videos and can do so at a temporal granularity higher than that available within the dataset. This quality makes our approach applicable to generic datasets. Further discussion of closely related works and the limitations of our method is provided in Appendix~\ref{appendix:additional_discussion}.

As generative video models embark on a journey toward genuine world simulators, ensuring that their output adheres to fundamental physical laws becomes paramount for their application in high-stakes domains such as robotics, healthcare, and autonomous systems. By introducing a method for localized, physics-aware conditioning, this work contributes a meaningful step toward that ambitious goal.

\section*{Acknowledgments}
This work is supported by the EU funded \emph{SYNERGIES} project (Grant Agreement No. \emph{101146542}). We also gratefully acknowledge the \emph{TUE} supercomputing team for providing the \emph{SPIKE-1} compute infrastructure to carry out the experiments reported in this paper.
{
    \small
    \bibliographystyle{ieeenat_fullname}
    \bibliography{main}
}

\clearpage
\setcounter{page}{1}
\maketitlesupplementary
\section{Additional Information}
\label{appendix:additional_discussion}
We begin with preliminaries, wherein we provide a brief overview of \ac{RoPE} followed by a procedural description of the chunk aware cross-attention mechanism to augment the discussion in~\cref{sec:method}. Subsequently, we supplement~\cref{sec:conclusion} with a discussion of works closely related to \paper followed by a discussion on its limitations and future opportunities for contribution.

\subsection{Rotary Positional Embeddings (RoPE)}
\ac{RoPE} is an established method for encoding positional information within transformer-based models that uniquely captures both absolute and relative positional data through vector rotations~\cite{su2024roformer}. Fundamentally, the goal is to apply a set of block-diagonal rotation matrices $R$ to the query vectors $q$ and the key vectors $k$ at each position. Thus, the transformations for the $d$ dimensional query vector $q_m$ and the key vector $k_n$ at the positions $m$ and $n$, respectively, are
\begin{align}
    q_m' = R_mq_m \qquad k_n'=R_nk_n
\end{align}
where $R_m$ and $R_n$ are the corresponding block-diagonal rotation matrices consisting of $d/2$ blocks. Each diagonal block $R_{m,i}$ in $R_m$ corresponds to dimensions $2i-1, 2i$ and is defined as
\begin{align}
R_{m,i} = \begin{pmatrix}
\cos(m\theta_i) & -\sin(m\theta_i)\\
\sin(m\theta_i) & \cos(m\theta_i)
\end{pmatrix}
\end{align}
where $\theta_i$ is a predefined frequency term.
Since $R_m^TR_n = R_{n-m}$, this design ensures that the inner product of the rotated query and key vectors $(q_m')^Tk_n'$ depends only on the original query and key vectors and their relative distance, $n-m$.

\subsection{Chunk aware cross attention procedure}
In \paper, the chunk-wise positional alignment between visual query tokens and textual key tokens within the local pathway is achieved through a coherent grid-based encoding scheme. The procedure is described in~\cref{alg:ropealign}. Specifically, the text tokens from all video segments are concatenated and introduced into the local attention pathways as illustrated in~\cref{fig:training}. To preserve and utilize the contextual position of each token, a two-dimensional coordinate grid is imposed on the set of text key tokens. Within this grid, the first dimension indexes the corresponding video chunk, while the second dimension identifies the intra-chunk position. \ac{RoPE} is then applied to encode these 2D coordinates in the representation of each key token. This design ensures that global and local positional information is preserved for each token throughout the network. Thus, the subsequent cross-attention mechanism can attend to localized content across all chunks, while maintaining precise chunk-specific referencing and temporal awareness.

\begin{algorithm}
    \caption{Chunk Aware Cross-Attention}
    \label{alg:ropealign}
    \begin{algorithmic}[1] % [1] for line numbers
        % ---------------------------
        % INPUTS / OUTPUTS
        % ---------------------------
        \Require Video tokens $X \in \mathbb{R}^{B \times L_v \times H \times d}$   \Comment{$L_v$: video sequence length}
        \Require Local text representations $\{T^{(b)}\}_{b=1}^{N_b}$ for $N_b$ video chunks
        \Require Video grid $G_v \in \mathbb{N}^{B \times 3}$, \ac{RoPE} frequencies $\Omega$
        \Require Number of chunks $N_b$, per-chunk text length $L_c$
        \Ensure Updated video representation $\hat{X}$ after chunk aware cross-attention

        % ---------------------------
        % Proocedure
        % ---------------------------
        \Statex \textbf{// 1. Concatenate local text across all chunks}
        \State $T \gets \mathrm{Concat}(T^{(1)}, T^{(2)}, \dots, T^{(N_b)})$ \Comment{Single sequence of length $L_t = N_b \cdot L_c$}
        \Statex
        
        \Statex \textbf{// 2. Build 2D grid over local text tokens. Initialize $G_t \in \mathbb{N}^{B \times 3}$ as follows:}
        \State $G_t[:,0] \gets N_b$ \Comment{first grid dimension = chunk index}
        \State $G_t[:,1] \gets L_c$ \Comment{second grid dimension = intra-chunk position}
        \State $G_t[:,2] \gets 1$ \Comment{dummy spatial axis to share ApplyRoPE API for both video and text tokens}
        \Statex
        
        \Statex \textbf{// 3. Compute query, key, and value representations}
        \State $Q \gets \mathrm{ProjectAndNormalize\_Video}(X, W_q)$    \Comment{Video queries}
        \State $K \gets \mathrm{ProjectAndNormalize\_Text}(T, W_k)$ \Comment{Local text keys}
        \State $V \gets \mathrm{Project\_Text}(T, W_v)$ \Comment{Local text values}
        \Statex
        
        \Statex \textbf{// 4. Apply \ac{RoPE} using video and text grids}
        \State $\widetilde{Q} \gets \mathrm{ApplyRoPE}(Q, G_v, \Omega)$ \Comment{Encode video tokens with (frame, height, width) positions}
        \State $\widetilde{K} \gets \mathrm{ApplyRoPE}(K, G_t, \Omega)$ \Comment{Encode text tokens with (chunk, intra-chunk) positions}
        \Statex
        
        \Statex \textbf{// 5. Multi-head cross-attention over all concatenated chunks}
        \State $\hat{X} \gets \mathrm{MultiHeadAttention}(\widetilde{Q}, \widetilde{K}, V)$ \Comment{Attend from each video token to all local text tokens across chunks}
        \State \Return $\hat{X}$  \Comment{Video features updated with chunk-aware local text information}
    \end{algorithmic}
\end{algorithm}

\subsection{Related work} The proposed work is in line with recent studies that address the limitations of cross-attention mechanisms within generative \acf{T2V} frameworks based on \acfp{DiT}.
A closely related concept is ``Segmented Cross-Attention'' introduced in Presto~\cite{yan2025long}, where a prompt is divided into sub-captions using a \ac{LLM}, each aligned to a specific temporal segment of the video. This method is a parameter-free mechanism for generating long-range videos that follow a sequence of narrative instructions derived from the main caption. Similarly, DiTCtrl~\cite{cai2025ditctrl} is a training-free method that enables multi-prompt video generation by controlling attention to create smooth transitions between different textual conditions over time. These methods aim to improve narrative coherence using explicit sub-prompts that are generated purely from text or are explicitly provided, while still relying on modulation of the global attention pathway. Although \paper also aligns textual information with local temporal segments, its objective and mechanism are distinct. In contrast to these methods, our method does not redesign or modulate the core attention module. Instead, we introduce new, separate cross-attention blocks as a modular addition to a pretrained model, specifically to integrate the chunk-wise generated physics prompts, thereby complementing the global prompt without affecting its attention pathways.

\subsection{Limitations and future scope} Although video-understanding capabilities in \acp{VLM} have improved significantly in recent years, they are still prone to hallucination and can produce information that is completely incorrect or misaligned with the visual content presented. This fundamental challenge currently limits their ability to reliably extract physics information from longer videos or videos with complex spatiotemporal physical content. Furthermore, annotating larger datasets with \ac{VLM} also requires an additional compute budget. Another challenge is scalability to larger models, since the model size can increase quickly due to additional layers in each transformer block. Therefore, an observed improvement in the physical awareness of the resulting model comes at the cost of slower training and inference over the corresponding baseline. However, this challenge can be mitigated to some extent with the help of advanced techniques for faster sampling, such as model distillation. A more theoretical limitation is the classic train-test distribution mismatch, since, during inference, \ac{VLM} does not have visual input to generate local annotations and must rely on global text alone. However, the video generator always sees the same interface, which is a sequence of local physics-aware text prompts. The mismatch therefore lies only in the upstream prompt generation bounded by the consistency with which the \ac{VLM} maps global descriptions to local physics statements with and without visual input. Our experiments on two benchmarks indicate that this does not prevent robust gains in the \acf{PC} score. Finally, as described in~\cref{sec:method}, while including global \ac{T2V} prompt in the instruction to the \ac{VLM} during the annotation of a video chunk helps generate annotations that are aligned with the global prompt, they do not explicitly prevent semantic misalignment of annotations among different video chunks. Future work could explore measures to reduce the computational cost of additional local pathways and improve alignment of locally extracted physics information across all chunks.

\section{VLM Instructions}
In this section, we provide the details on the instructions given to the \ac{VLM} for different use cases.

\label{appendix:vlmprompts}

\subsection{Physics grounded video chunk annotation}
\begin{figure*}
  \centering
  \includegraphics[width=\textwidth]{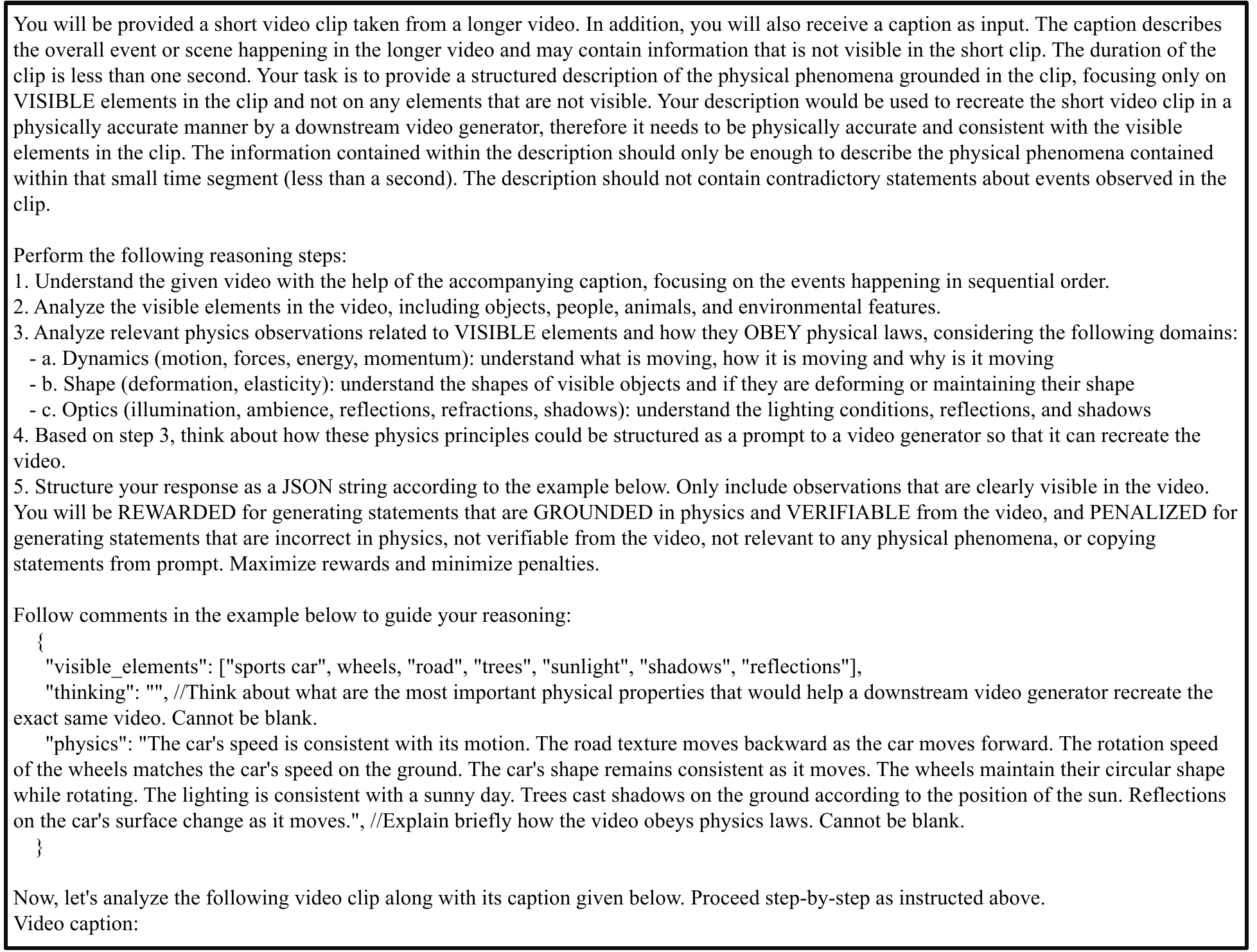}
    \caption{VLM instruction to generate the physics caption for a video chunk}
    \label{fig:positive_prompt}
\end{figure*}

\Cref{fig:positive_prompt} shows the \ac{VLM} input instruction used to generate physics grounded annotations for video chunks prior to training. During annotation, the global \ac{T2V} caption is appended to the \ac{VLM} instruction along with a contiguous chunk of frames from the input video, as discussed in~\cref{sec:method}.
\label{appendix:posprompt}

\subsection{Counterfactual annotation}
\begin{figure*}
  \centering
  \includegraphics[width=\textwidth]{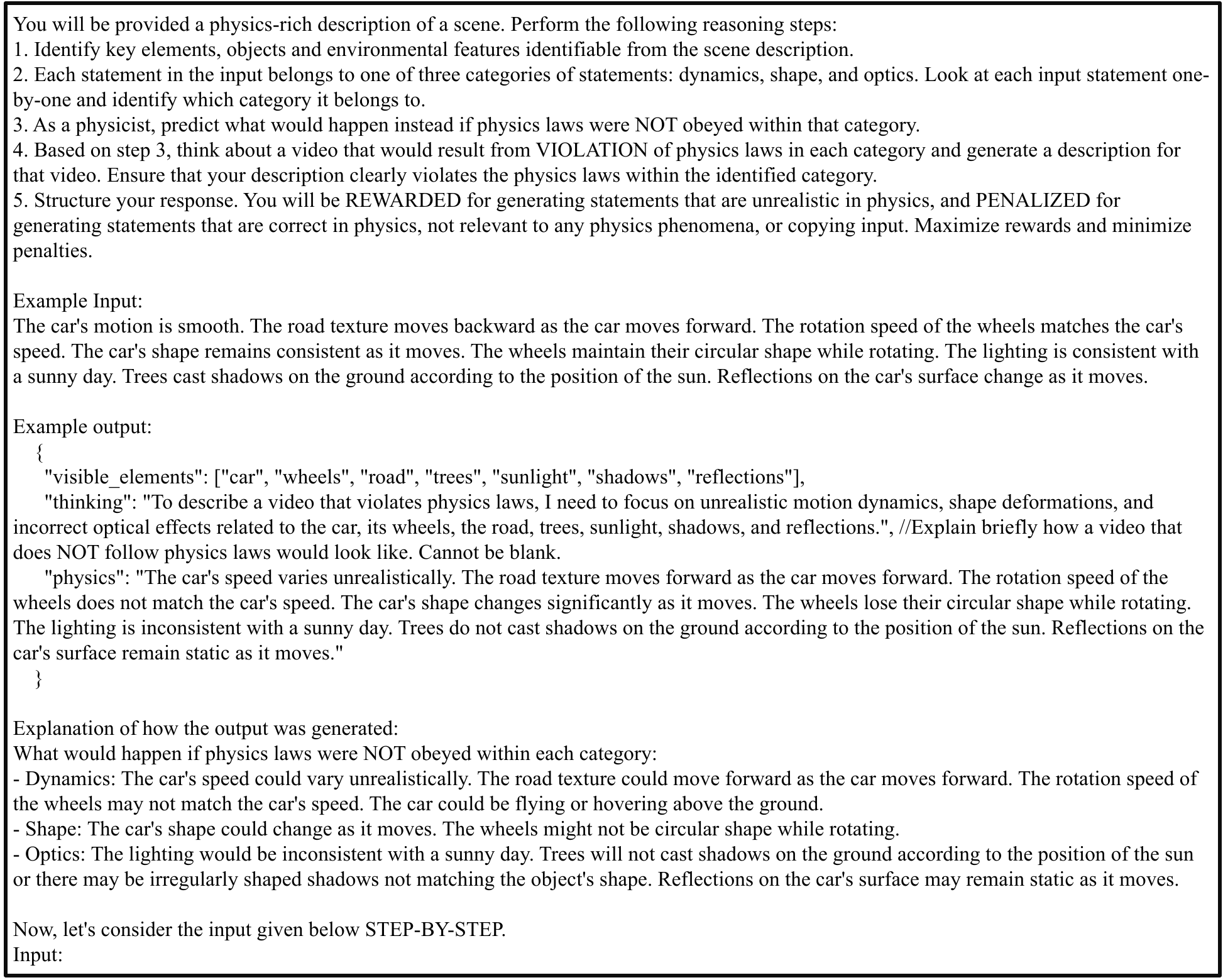}
    \caption{VLM instruction to generate the counterfactual physics caption}
    \label{fig:negative_prompt}
\end{figure*}

\Cref{fig:negative_prompt} shows the \ac{VLM} input instruction used to generate the counterfactual prompt based on incorrect physics. The counterfactual annotation in our method relies only on the generated ``positive'' local prompt for a given chunk and does not use any other information. As discussed in~\cref{sec:method}, this helps prevent the generation of physically correct descriptions which are undesirable in this phase.
\label{appendix:negprompt}

\subsection{Physics-grounded local prompt generation during inference}
\begin{figure*}
  \centering
  \includegraphics[width=\textwidth]{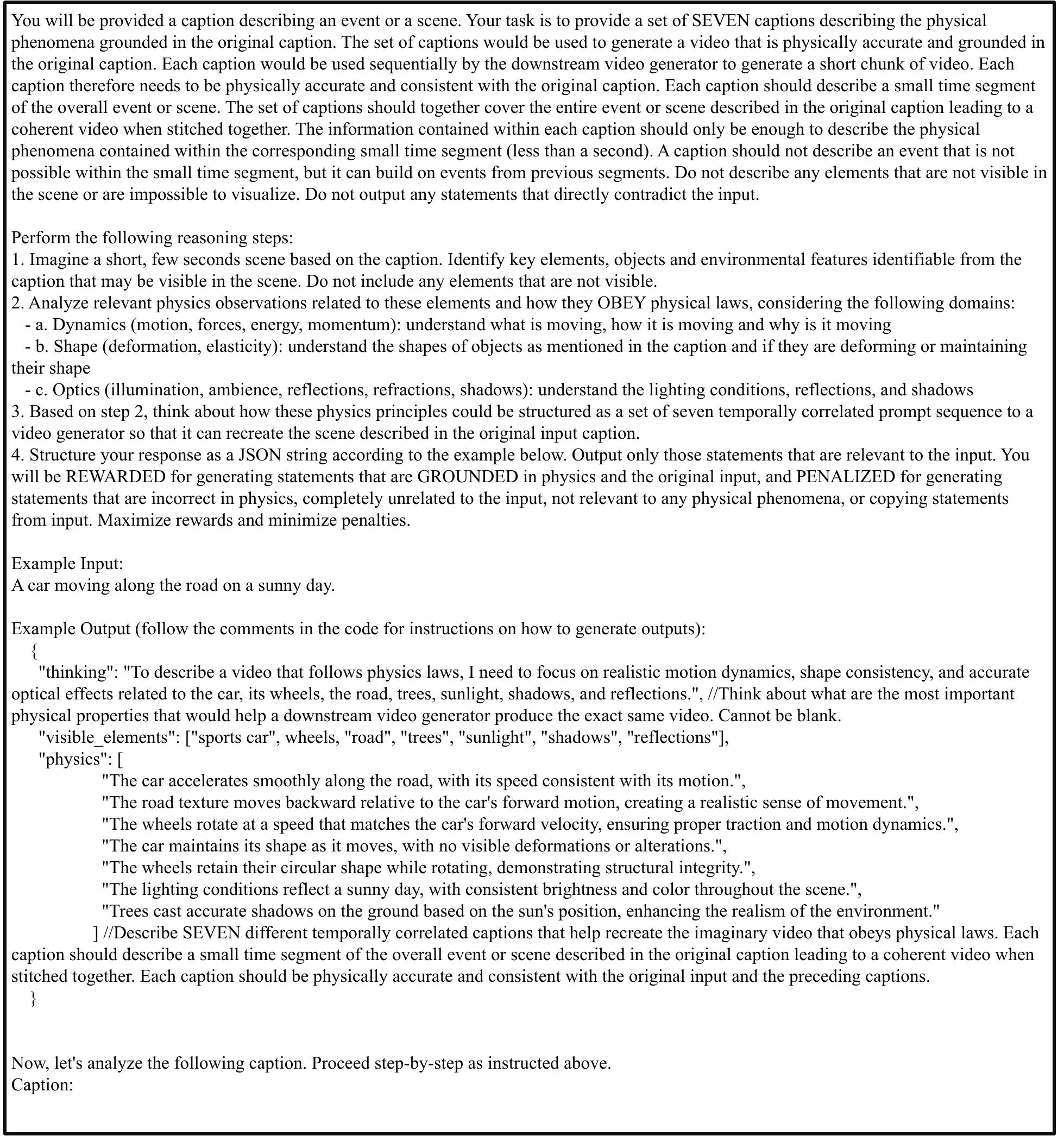}
    \caption{VLM instruction to generate the local physics captions for all the chunks to be generated at once during inference}
    \label{fig:inference_prompt}
\end{figure*}

During inference, the visual data is not available. However, we still need the local physics based instructions to be provided as input to the model. To ensure this, we use the \ac{VLM} instruction shown in~\cref{fig:inference_prompt}. This instruction relies only on the information contained in the global \ac{T2V} caption to generate a coherent set of physically correct local prompts.
\label{appendix:infprompt}

\section{Annotation Examples}
\label{appendix:example_prompts}
\begin{figure*}
  \centering
  \includegraphics[width=\textwidth]{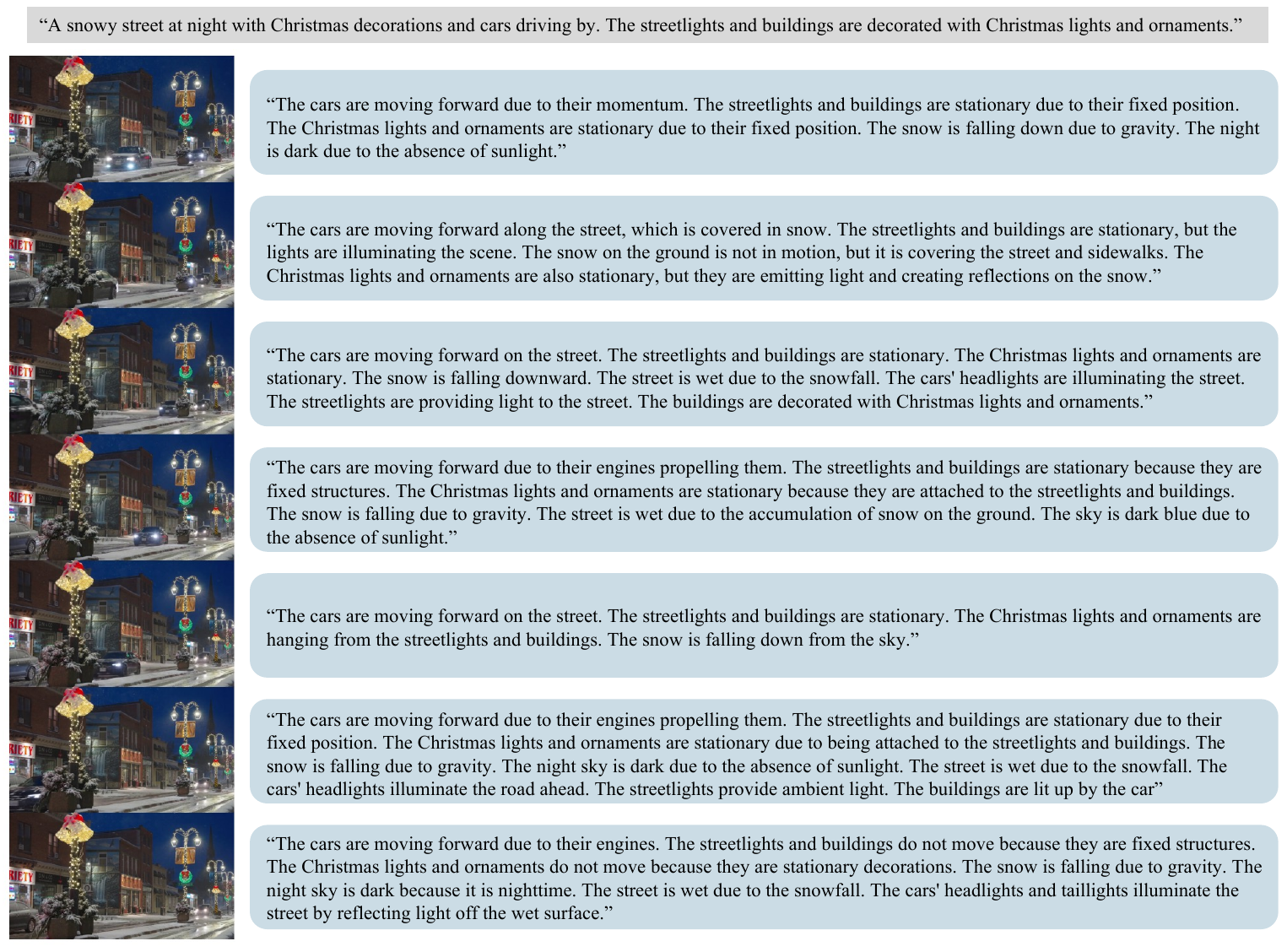}
    \caption{A sample from the data with annotations generated by a \acf{VLM}. The topmost text is the global prompt and the local annotations are listed sequentially alongside a representative frame from each chunk.}
    \label{fig:annotations}
\end{figure*}

\begin{figure*}
  \centering
  \includegraphics[width=\textwidth]{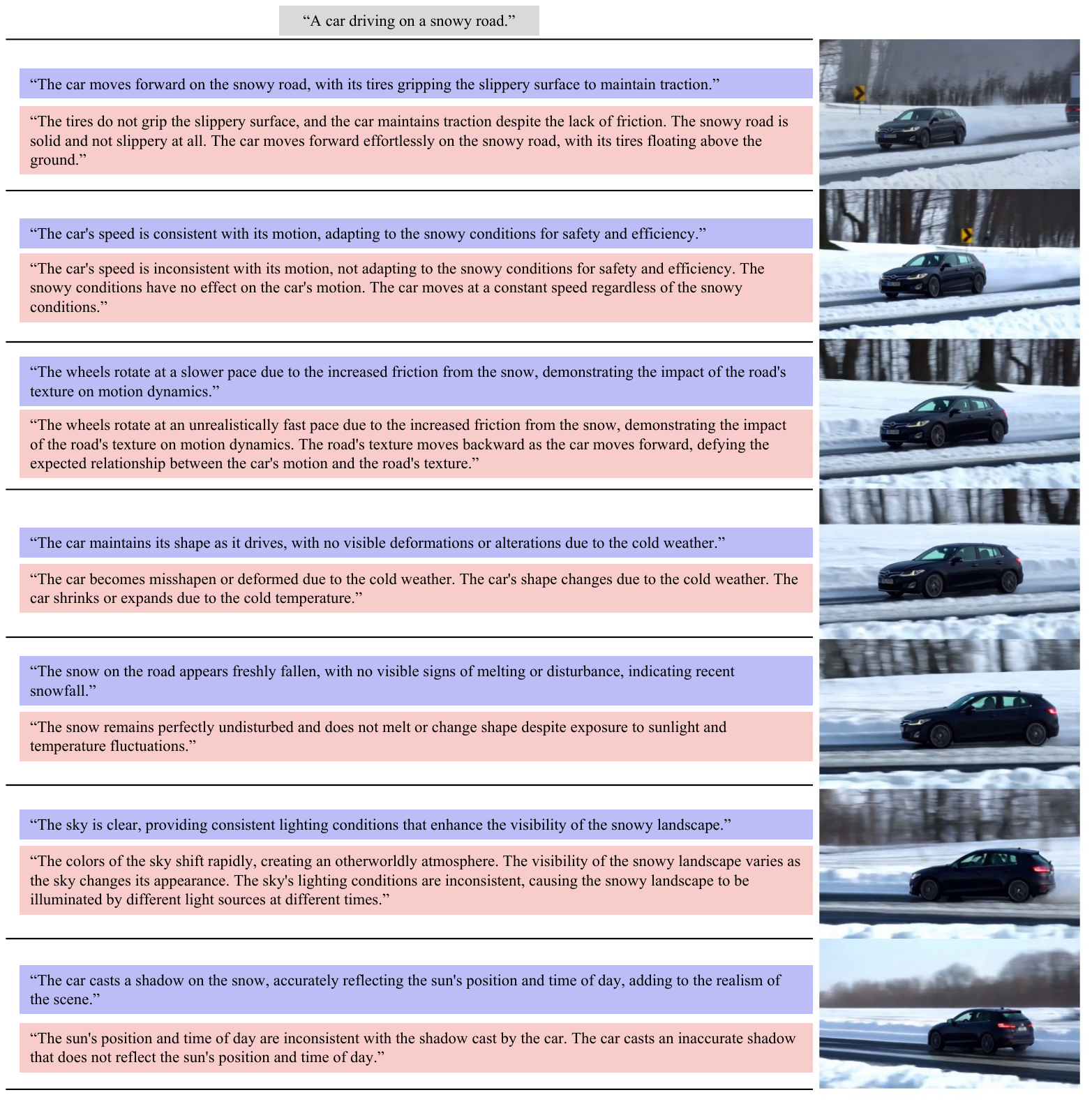}
    \caption{A set of physics\colorbox{mypurple}{grounded} and physics\colorbox{myred}{counterfactual} prompts generated during inference. The representative frames from the generated video chunks are shown on the right.}
    \label{fig:inference_annotations}
\end{figure*}

In~\cref{fig:annotations}, we visualize the global and local annotations generated by \ac{VLM} for an example in the training data set, together with the representative frames for each chunk. Similarly, during inference,~\cref{fig:inference_annotations} we provide the local annotations generated by the \ac{VLM} for an example caption, along with the representative frames from the video chunks generated using these annotations.

\section{Additional Results}
\subsection{Qualitative examples}
\label{appendix:additional_qualitative}
\begin{figure*}
  \centering
  \includegraphics[width=.65\textwidth,trim=4 8 4 4,clip]{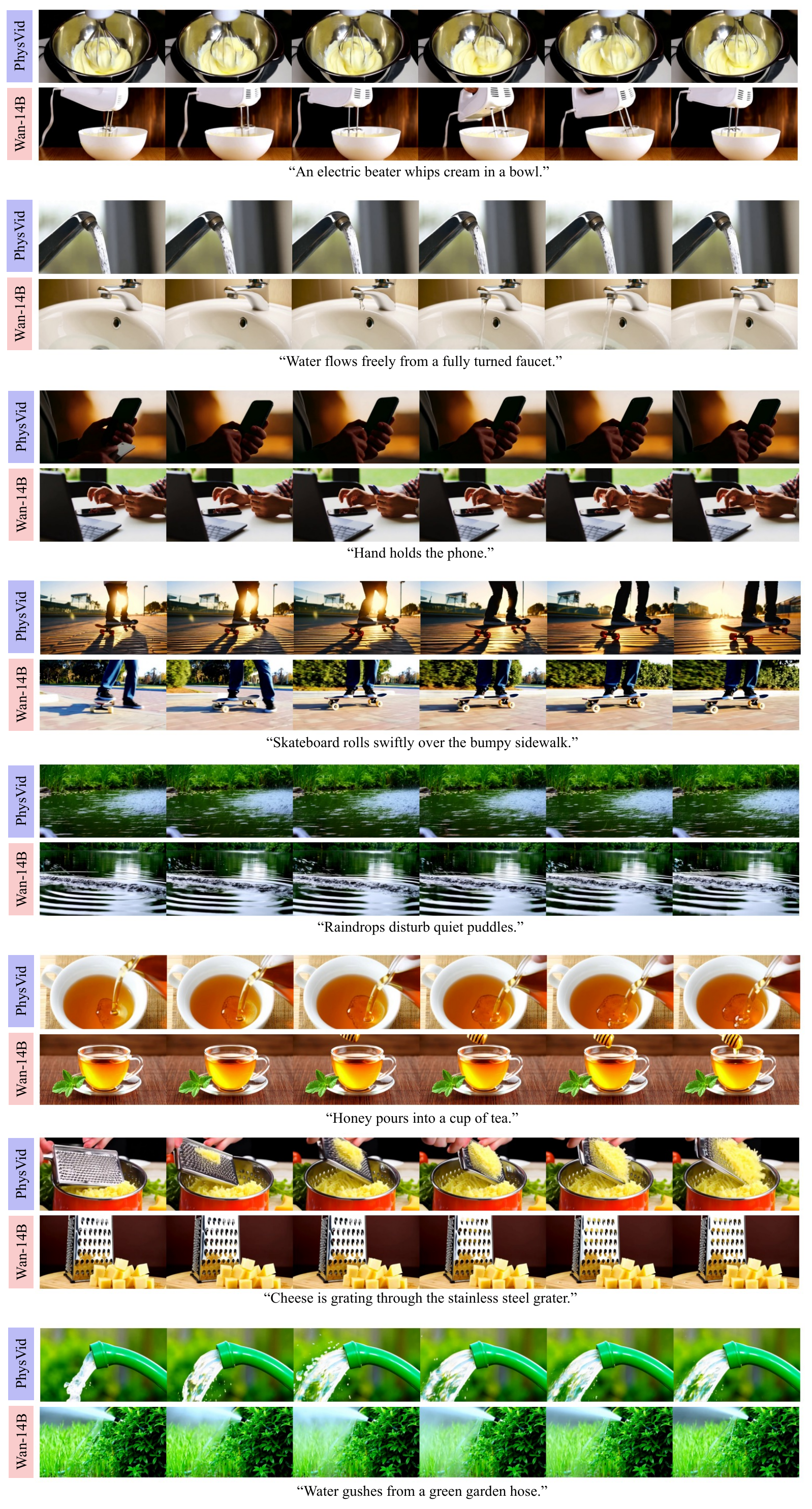}
    \caption{Additional comparisons between \emph{\paper} and \emph{Wan-14B}. Captions are from VideoPhy.}
    \label{fig:more_qualitative}
\end{figure*}

\begin{figure*}
  \centering
  \includegraphics[width=0.78\textwidth]{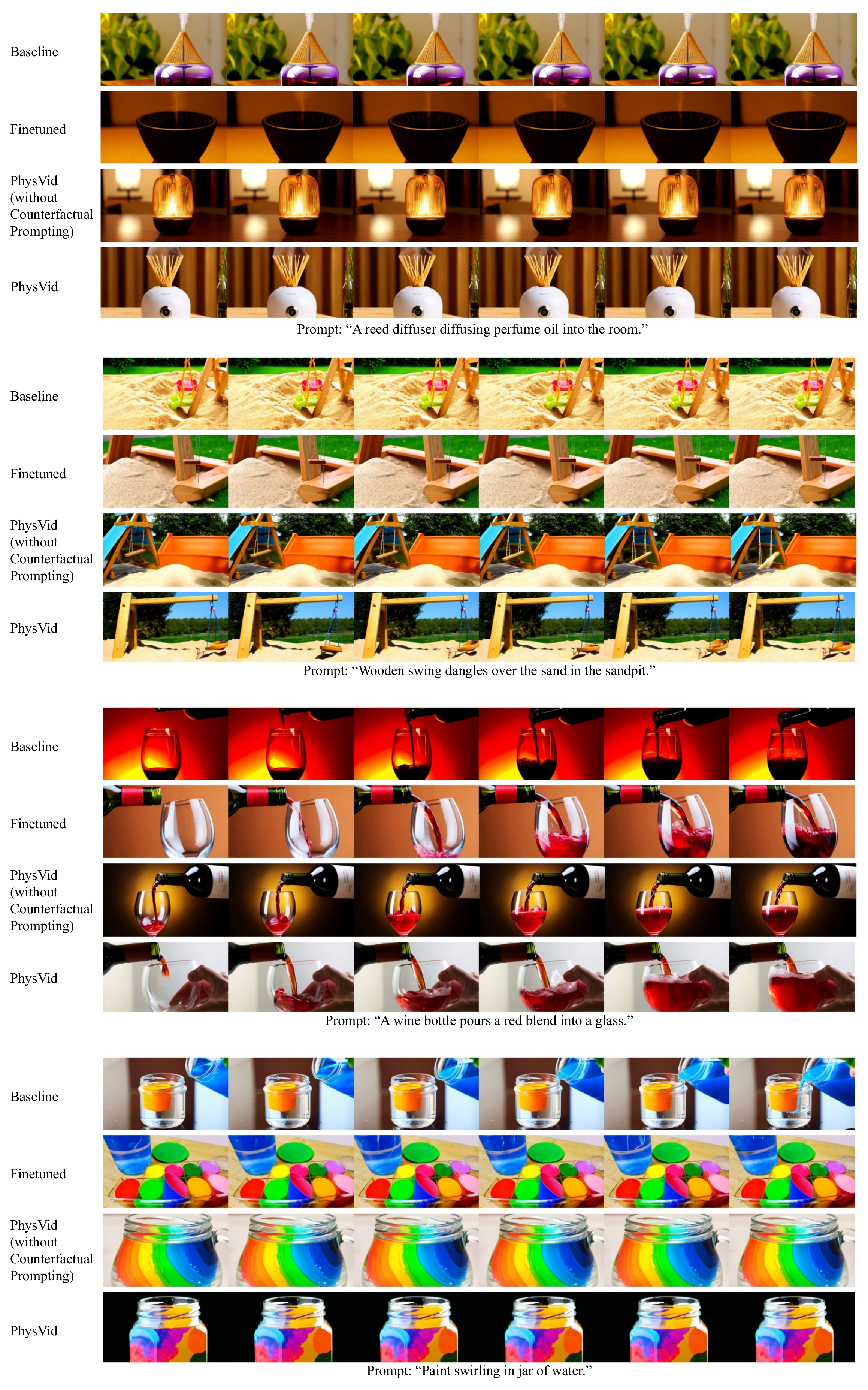}
    \caption{Supplementary qualitative results from the ablation study. All prompts are from VideoPhy.}
    \label{fig:more_ablations}
\end{figure*}

To supplement the examples in~\cref{fig:main}, we visualize additional results generated by our method in~\cref{fig:more_qualitative} with comparisons to Wan-14B. In~\cref{fig:more_ablations}, we also provide additional qualitative results from the ablation study discussed in~\cref{sec:ablations}, to supplement~\cref{fig:qualitative_ablation}.
These examples are also available as videos along with additional video examples on our \href{https://5aurabhpathak.github.io/PhysVid}{project website}.

\subsection{Similarity metrics}
We evaluated \paper on \emph{four} similarity metrics as shown in~\cref{tab:perceptual}. As can be observed, the results remain consistent, showing a slightly increased FVD score relative to the finetuned baseline, which corroborates the previously noted minor compromise in content fidelity (see~\cref{tab:videophy12}) in exchange for substantial improvements in physical realism.

\begin{table}[h]
    \centering
    \begin{tabular}{@{}lcccl@{}}
         \hline
         Model & LPIPS$\downarrow$ & FVD$\downarrow$ & SSIM$\uparrow$ & PSNR$\uparrow$\\
         \hline
         Wan 1.3B & 0.703& 417.352& 0.217& 8.625\\
         Wan (finetuned) & 0.671& 302.465& 0.239& 9.379\\
         \paper & 0.679& 318.087& 0.240& 9.234\\
         \hline
    \end{tabular}
    \caption{Additional metrics based on similarity (2048 sample pairs)}
    \label{tab:perceptual}
\end{table}

\section{Other details}
\begin{table*}
    \centering
    \caption{Configurations}
    \begin{tabular}{lccl}
        \toprule
        \multicolumn{2}{c}{Training}\\
        \midrule
        Base architecture &  Wan-1.3B\\
        Additional parameters (M) &  $400$\\
        Effective batch size &  $64$\\
        Number of steps & $3000$\\
        Number of epochs & $4$\\
        Learning rate (Stage 1: 1000 steps, frozen base layers) & $1\times10^{-5}$\\
        Learning rate (Stage 2: 2000 steps, full architecture) & $2\times10^{-6}$\\
        Loss & Flow Matching\\
        Optimizer & AdamW\\
        Timestep Shift Factor & $8$\\
        Number of Latent Frames per Chunk & $3$\\
        Number of Latent Chunks & $7$\\
        \midrule
        \multicolumn{2}{c}{Inference}\\
        \midrule
        Number of Denoising Steps & $50$\\
        Guidance Scale & $6$\\
        Wan-1.3B Latency per Video (s) & $66$\\
        Wan-14B Latency per Video (s) & $310$\\
        \paper Latency per Video (s) & $93$\\
        \midrule
        \multicolumn{2}{c}{Video}\\
        \midrule
        Resolution & $832\times480$\\
        FPS & $16$\\
        Duration (s) & $5.06$\\
        Frames & $81$\\
        \bottomrule
    \end{tabular}
    \label{tab:other_details}
\end{table*}

\paragraph{Configuration}~\Cref{tab:other_details} lists hyperparameter configurations and other relevant settings for training and inference.

\paragraph{VLM overhead} The average VLM overhead for all prompts generated per sample is $\approx 17.22\pm0.99$ seconds. Including the denoising loop runtime quantified in~\cref{tab:other_details}, overall, \paper approach takes $110$ seconds per video, which is near third ($0.35\times$) of $310$ seconds per video latency of Wan-14B. All inference run times in our work are reported with \textit{bfloat16} precision on a single B200 GPU with a batch size of 1.

\end{document}